\documentclass[letterpaper, 10 pt, conference]{ieeeconf}  

\IEEEoverridecommandlockouts                              

\usepackage{cite}
\usepackage{multirow}
\usepackage{graphicx}
\usepackage{subfigure}
\usepackage{caption}
\usepackage{amsmath}
\usepackage{amsfonts,amssymb}
\usepackage{algorithm}
\usepackage{algorithmic}
\usepackage{color}
\usepackage{xcolor}
\usepackage[colorlinks,bookmarksopen,bookmarksnumbered,citecolor=green, linkcolor=red, urlcolor=blue]{hyperref}

\title{\LARGE \bf
SA-Attack: Speed-adaptive stealthy adversarial attack on trajectory prediction
}

\author{Huilin Yin$^{1}$, Jiaxiang Li$^{1}$, Pengju Zhen$^{1}$ and Jun Yan$^{1}$
\thanks{$^{1}$Huilin Yin, Jiaxiang Li, Pengju Zhen and Jun Yan (yinhuilin, 2230747, 2230750, yanjun@tongji.edu.cn) are with the College of Electronics and Information Engineering, Tongji University, Shanghai 201804, China}%
}
\begin{document}

\maketitle
\thispagestyle{empty}
\pagestyle{empty}

\begin{abstract}

Trajectory prediction is critical for the safe planning and navigation of automated vehicles. The trajectory prediction models based on the neural networks are vulnerable to adversarial attacks. Previous attack methods have achieved high attack success rates but overlook the adaptability to realistic scenarios and the concealment of the deceits. To address this problem, we propose a speed-adaptive stealthy adversarial attack method named SA-Attack. This method searches the sensitive region of trajectory prediction models and generates the adversarial trajectories by using the vehicle-following
method and incorporating information about forthcoming trajectories. Our method has the ability to adapt to different speed scenarios by reconstructing the trajectory from scratch. Fusing future trajectory trends and curvature constraints can guarantee the smoothness of adversarial trajectories, further ensuring the stealthiness of attacks. The empirical study on the datasets of nuScenes and Apolloscape demonstrates the attack performance of our proposed method. Finally, we also demonstrate the adaptability and stealthiness of SA-Attack for different speed scenarios. Our code is available at the repository: \url{https://github.com/eclipse-bot/SA-Attack}.
\end{abstract}

\section{INTRODUCTION}

Automated vehicles (AVs) integrate a variety of modules, including environmental perception, driving behavior planning, and vehicle control. Trajectory prediction is a crucial component of autonomous driving, which predicts the future trajectory of nearby moving objects for the planning and navigation. Recent data-driven trajectory prediction methods
have shown remarkable performances on motion forecasting benchmarks\cite{yuan2021agentformer,yin2021memory,salzmann2020trajectron++,li2019grip++}. However, these trajectory prediction models based on deep learning are vulnerable to adversarial attacks. Therefore, it is necessary to investigate the robustness of trajectory prediction models under the adversarial attacks.


The key challenge in adversarial attacks on trajectory prediction is how to generate effective attacks while guaranteeing their stealthiness. 
On the one hand, the adversarial attack should be powerful enough to deceive the trajectory prediction model, leading it to make unsafe or unreasonable prediction. Thus, the adversarial perturbations should be restricted to the bounded sensitive regions of trajectories. Previous studies have paid little attention to this analysis process. To best of our knowledge, these studies\cite{cao2022advdo,zhang2022adversarial} obtain malicious perturbations by searching directly under the constraints. It makes the results prone to the local extremum with a high computational cost. 


On the other hand, trajectories generated by the adversarial attacks should ensure naturality~\cite{wang2021human} so that they are not easily recognised as anomalous trajectories by the AV system. First, the adversarial trajectory must adhere to physical constraints, including dynamics and kinematics constraints\cite{pivtoraiko2011kinodynamic,sprunk2011online,andreasson2015fast}, to guarantee its feasibility in the real world. Second, the adversarial trajectory should exhibit smoothness, which helps minimize abnormal driving behaviors such as sharp turns or jerks. Finally, the adversarial trajectory should maintain a similar intent\cite{zhao2021tnt} to the original trajectory to ensure that the original target is accessible. Previous works~\cite{zhang2022adversarial} have relied on statistical methods to derive average values of specific physical properties, such as speed and acceleration. However, employing statistical data to enforce constraints is inadequate for addressing complex, high-dimensional scenarios. 
Furthermore, to the best of our knowledge, these studies paid little attention to future trajectory information during the generation of adversarial trajectory. This may result in a change in the intent of the original trajectory, causing anomalies in the trajectory junctions.


To address these issues, we propose an effective speed-adaptive adversarial trajectory generation method named SA-Attack. We first search without constraints to obtain trajectory shapes that are sensitive to the trajectory prediction model. Subsequently, we reconstruct the feasible trajectories using vehicle-following method. These trajectories will be naturally biased towards model-sensitive trajectories under feasible conditions. Finally, to ensure smooth transitions between the adversarial trajectories and future trajectories, we incorporate information about forthcoming trajectories. Our proposed SA-Attack method achieves considerable attack performance on the nuScenes~\cite{caesar2020nuscenes} and Apolloscape~\cite{huang2018apolloscape} datasets.

Our main contributions are summarized as follows:

\begin{itemize}
\item We propose a novel adversarial trajectory generation method which starts by determining the trajectory shapes sensitive to the trajectory prediction model through an unconstrained search. The feasible trajectories are later reconstructed using vehicle-following method. Our method has the ability to adapt to different speed scenarios and generate stealthy adversarial trajectories. The extensive experiment validates the effectiveness of our proposed method.
\item We use a continuous curvature model in the generation of adversarial trajectories. By using a continuous curvature model rather than a dynamic model, we can describe the geometrical properties of the motion trajectories and enhance the smoothness of the generated trajectories.
\item During the generation of the adversarial trajectory, we combine the future trajectory information to capture the intent of the original trajectory, ensuring the smoothness of the junction between the adversarial trajectory and the future trajectory.
\end{itemize}

\section{RELATED WORKS}

\subsection{\textbf{Deep-learning-based trajectory prediction}}

Due to the ability to handle complex spatio-temporal dependencies and adapt to multimodal data, deep-learning-based trajectory prediction models become the focus of research in recent years. These models\cite{salzmann2020trajectron++,li2019grip++,yuan2021agentformer,yin2021memory} use the spatial coordinate position of the trajectory as the primary input and incorporate other features e.g., interaction between agents, map information, and dynamic model to improve the prediction accuracy of the model. 

Trajectron++~\cite{salzmann2020trajectron++} is one of the representative approaches, which models temporal features and interaction of agents through recurrent neural networks\cite{alahi2016social,vemula2018social,yan2020trajectory} and a graph neural network\cite{gao2020vectornet,zhou2022hivt}, respectively. The approach takes into account both the dynamics constraints of the agents and other heterogeneous data. The model demonstrates the ability to handle multimodal data on the nuScenes dataset\cite{caesar2020nuscenes}.

The other representative approach is Grip++~\cite{li2019grip++}, which similarly constructs a graph to represent the interaction between agents and converts other information about the trajectories into a specific format for subsequent efficient computation. The model captures features in temporal and spatial terms by alternately using graph operations\cite{zhang2019graph} and temporal convolution\cite{cui2019multimodal}. The model achieves significant prediction results on the ApolloScape dataset\cite{ma2019trafficpredict}.

For security reasons, we still need to be concerned about the robustness of trajectory prediction models despite the high accuracy they achieve on different datasets. In particular, whether these models can maintain their original prediction accuracy under the malicious adversarial attacks is an important issue.

\subsection{\textbf{Adversarial attack on trajectory prediction}}
In AV systems, some modules based on deep learning such as object detection\cite{zhang2019towards,yan2023adversarial}, object tracking\cite{jia2020fooling}, lane detection\cite{sato2021dirty} can be affected by adversarial attacks and create security issues. Recent years, adversarial attacks on trajectory prediction models have received a lot of attention. Zhang et al.~\cite{zhang2022adversarial} first propose an adversarial attack method for trajectory prediction, in which they generate adversarial trajectories by directly applying perturbations to the spatial positions of the trajectories, and then the optimal perturbations are searched under a given constraint. However, the constraint using a fixed value cannot be adapted to different speed scenarios. Cao et al.\cite{cao2022advdo} improve on this method by applying perturbations to control signals such as acceleration and curvature, after which adversarial trajectories are recovered by a dynamic model. However, the adversarial trajectory generated by this method gradually deviates from the original trajectory, causing anomalies at the junction of the adversarial trajectory and the future trajectory.

To this front, we propose an attack strategy that incorporates the vehicle-following approach. The adversarial trajectories are generated from scratch enabling them to adapt to different speed scenarios. Future trajectory information is considered to ensure that the trajectory junctions are natural. Our attack is stealthy and can be more easily reproduced in real scenarios.

\begin{figure*}[!t]
\centering
\includegraphics[width=1.0\textwidth]{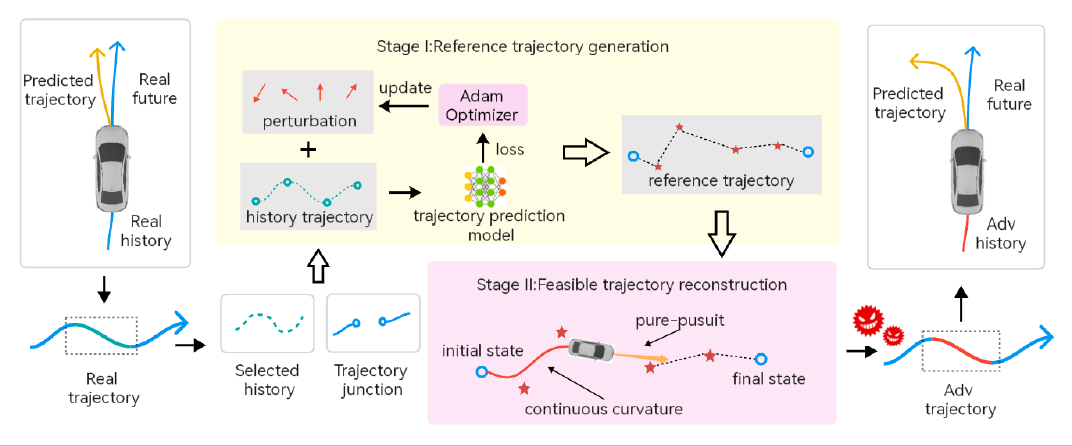}
\caption{Speed-adaptive stealthy adversarial attack (SA-Attack) methodology overview}
\label{fig:method}
\end{figure*}

\section{PROBLEM FORMULATION}

\textbf{Trajectory Prediction Formulation:} In this work, we focus on deep-learning-based trajectory prediction tasks. The goal of this task is to predict the distribution of future trajectories of $N$ agents in the scenario. Each agent has a semantic class, e.g. Car, Bus, or Pedestrian. We denote the state of agent as $s\in \mathbb{R}^{D}$. At time t, given the state of each agent and all of their histories for the previous $L_{I}-1$ timesteps, which we denote as $\mathbf{H}_{t}$, $\mathbf{H}_{t}=s_{t-L_{I}+1:t}^{1,...,N} \in \mathbb{R}^{L_{I}\times N\times D}$, as well as other available information for each agent $\mathbf{I}^{1,...,N}_{t}$, we seek a distribution over all agents’ future states for the next $L_{O}$ timesteps $\mathbf{P}_{t}=s_{t+1:t+L_{O}}^{1,...,N}\in \mathbb{R}^{L_{O}\times N\times D}$, which we denote as $\rho (\mathbf{P}_{t}|\mathbf{H}_{t},\mathbf{I}_{t})$. We also denote the ground truth states of all agents in the next $L_{O}$ timesteps as $\mathbf{F}_{t}=f_{t+1:t+L_{O}}^{1,...,N}$.

\textbf{Adversarial Attack Formulation:} In this work, we focus on one vehicle close to the AV, which is called adversary vehicle. AV's trajectory prediction model predicts the future trajectory of adversary vehicle and plans accordingly. Adversary vehicle maximise prediction error by driving along a crafted trajectory, which causes AV to perform unsafe driving behaviour. In this work, we focus on white-box attack methods, where the adversary vehicle has access to model parameters, history state $\mathbf{H}_{t}$ and future state $\mathbf{F}_{t}$ of all agents. White-box attack methods\cite{yin2022adversarial} can leverage internal information about trajectory prediction model to conduct more precise and effective attacks in order to explore what a powerful adversary can do. According to the Kerckhoffs’s principle\cite{shannon1949communication}, white-box attacks can motivate more effective defense methods.


\section{METHOD}

Our proposed SA-Attack method is visualized in Fig.\ref{fig:method}. Specifically, our approach consists of two stages: (1) Reference trajectory generation, and (2) Feasible trajectory reconstruction. In the first stage, we find model-sensitive trajectory shape by modifying the original trajectory in the lane range. After that, we concatenate the obtained sensitive trajectory points with the real trajectory junction to obtain the reference trajectory. In the second stage, we leverage a vehicle-following approach and continuous curvature model to simulate vehicle driving and generate feasible adversarial trajectory, which can mislead future trajectory prediction.

\subsection{Reference trajectory generation}

First, we generate multiple sets of randomly initialized perturbations in order to cover as many trajectory shapes as possible. Next, we add each set of perturbations to the history states $s_{t-L_{I}+1:t}$, which only change the spatial coordinate position of the trajectory to explore model-sensitive trajectory points. Finally, we use a white-box optimisation method based on Projected Gradient Descent (PGD)\cite{madry2017towards} to update the perturbations in a small space. In each iteration, we constrain each perturbation with a maximum module length of $1m$. As mentioned in the previous work~\cite{zhang2022adversarial}, the $1m$ deviation is an upper bound for a car not shifting to another lane if it is normally driving in the center of the lane.

In order to guarantee the smoothness of the junction between the adversarial trajectory and the real trajectory, we choose the real state as the initial and final state of the subsequent feasible trajectory reconstruction. We choose the previous state of the selected history trajectory as the initial state, or the first state of the history trajectory as the initial state if the selected history trajectory has no previous state. The first state of the future trajectory is used as the final state. Afterwards, the reference trajectory is obtained by concatenating the initial and final states with the history states after adding the optimal perturbations.

Note that the reference trajectory is used to describe the attack strategy from the adversaries' viewpoints, so it may not be a feasible trajectory. In addition, our attack goal is to guide the model to produce incorrect predictions that deviate from the real trajectory, so the perturbations are selected and optimised based on the root mean square error between the predicted trajectory and the future trajectory. 

\begin{algorithm}[!t]
\small
    \caption{Feasible trajectory reconstruction}
    \label{alg1}
    \begin{algorithmic}[1]
    \renewcommand{\algorithmicrequire}{\textbf{Input:}}
    \renewcommand{\algorithmicensure}{\textbf{Output:}}
    \ENSURE{feasible adversarial trajectory $\bar{s}_{t-L_{I}+1:t}$}
    \REQUIRE{reference trajectory, average speed $v$}
    \STATE{Set the initial and final configuration $p_{0},p_{n}$;}
    \STATE{Current step $i\gets 0$;}
    \WHILE{$p_{i}\ne p_{n}$}
    \STATE{Get the perceptual distance $p_{v}=\alpha \times v$;}
    \STATE{Get the desired curvature $c_{0,\mathrm{desired}}$;}
    \STATE{$c_{1,\mathrm{limited}}\gets FeasibleCurvatureRate(c_{0,i},c_{0,\mathrm{desired}},v)$;}
    \STATE{$A_{i+1}\gets (c_{1,\mathrm{limited}},l_{i+1})$;}
    \STATE{$p_{i+1}\gets ContinuousCurvatureModel(p_{i},A_{i+1})$;}
    \STATE{$i\gets i+1$;}
    \ENDWHILE
    \STATE{$\bar{s}_{t-L_{I}+1:t}\gets Sample(p_{0},...,p_{n})$;}
    \RETURN $\bar{s}_{t-L_{I}+1:t}$
    \end{algorithmic}
\end{algorithm}

\subsection{Feasible trajectory reconstruction} 

In this subsection, the feasible adversarial trajectory is reconstructed based on the reference trajectory, as shown in Algorithm \ref{alg1}. The algorithm delineates the method by which the continuous curvature model characterizes trajectories, as well as the process through which an adversary reconstructs these trajectories utilizing the pure-pursuit method.

\textbf{Continuous curvature model:} We choose part of the vehicle's state $s$ as the configuration $p$. The continuous curvature model computes the next configuration $p_{i+1}= \left(x_{i+1}, y_{i+1}, \psi _{i+1}, c_{0,i+1}\right)$ given the current configuration $p_{i}= \left(x_{i}, y_{i}, \psi _{i}, c_{0,i}\right)$ and clothoid arcs $A_{i+1}= \left(c_{1,i+1}, l_{i+1}\right)$, $i\in \{0,...,n-1\}$. Here, $x_{i},y_{i}, \psi_{i}, c_{0,i}$ represent 2D position, heading and curvature correspondingly, $c_{1,i+1}$ denotes $A_{i+1}$’s rate of change of curvature with respect to distance (curvature rate), and $l_{i+1}$ is its length.

Each arc $A_{i+1}$ induces a configuration $p_{i+1}$. Given the initial configuration $p_{0}$ and a sequence of clothoid arcs $A=(A_{1},...,A_{n})$, the other configurations are calculated recursively, based on the previous configuration $p_{i}$. Using slightly ambiguous notation, let $c_{0,i+1}(l)$ denote $A_{i+1}$'s curvature at length $l$, with $ 0\le l\le l_{i+1}$. For the continuous curvature model, arc's curvature varies linearly with its length. Consequently, it satisfies the formulation: $c_{0,i+1}(l)=c_{0,i}+c_{1,i+1}\times l$ and $c_{0,i+1}=c_{0,i+1}(l_{i+1})$. The heading at length $l$ is obtained by integrating the curvature along the arc defined in Eq. (\ref{eq:curvature}):
\begin{equation}\label{eq:curvature}
\psi _{i+1}(l)=\psi _{i}+\int_{0}^{l}c_{0,i+1}(\sigma )d\sigma ,
\end{equation}
with $\psi _{i+1}=\psi _{i+1}(l_{i+1})$. Finally, the Cartesian coordinates $(x_{i+1}(l),y_{i+1}(l))$ at $l$ are obtained by Eq. (\ref{eq:arc1}) and Eq. (\ref{eq:arc2}):
\begin{equation}\label{eq:arc1}
x_{i+1}(l)=x_{i}+\int_{0}^{l}\cos (\psi _{i+1}(\sigma ))d\sigma , 
\end{equation}
\begin{equation}\label{eq:arc2}
y_{i+1}(l)=y_{i}+\int_{0}^{l}\sin (\psi _{i+1}(\sigma ))d\sigma , 
\end{equation}
with $x_{i+1}=x_{i+1}(l_{i+1})$ and $y_{i+1}=y_{i+1}(l_{i+1})$.

As a result of the properties mentioned above, our trajectory's curvature is continuously changing, which guarantees the smoothness and physical feasibility of the generated trajectories. Smooth trajectories are more consistent with normal driving behaviour, reducing the risk of being detected as anomalous trajectories, and increasing the probability of success of the attack and the stealthiness of the attack.

\textbf{Pure-Pursuit reconstruction trajectory:} 

The reference trajectory comprises a collection of trajectory points that are sensitive to the model, aiming to maximize the prediction error of the trajectory prediction model. Subsequently, we use the pure-pursuit method\cite{snider2009automatic} to iteratively generate feasible trajectories, starting from the initial state and progressing along the reference trajectory towards the final state.

At each step $i$, we first calculate a speed-dependent perceptual distance $p_{v}$, a desired curvature $c_{0,\mathrm{desired}}$ is calculated based on $p_{v}$ defined in Eq. (\ref{eq:pv}). 
\begin{equation}\label{eq:pv}
c_{0,\mathrm{desired}}=\frac{2\times \vartheta }{p_{v}^2}
\end{equation}
$\vartheta$ is the lateral distance between the heading vector and the perceptual vector. Then, the vehicle model in \cite{fassbender2016motion} is applied. It generates a feasible curvature rate $c_{1,\mathrm{limited}}$ according to $(c_{0,i},c_{0,\mathrm{desired}},v)$. Finally, using the
current curvature $c_{0,i}$, the curvature rate $c_{1,\mathrm{limited}}$ and a fixed step size $0.2m$, a new clothoid arc is attached to the end of the trajectory. We repeat this process until the trajectory reaches the final state. However, the above method needs small steps to accurately recover model-sensitive trajectory shape, so the resulting trajectories contain excessive numbers of clothoid arcs. Next, we sample configurations $p_{0},...,p_{n}$ at equal intervals to obtain adversarial trajectories $\bar{s}_{t-L_{I}+1:t}$, which can be used as the input to the trajectory prediction model.

In the whole process of trajectory generation, we constrain the maximum curvature rate to simulate the maximum steering angle constraints in real applications. At the same time, the approach of reconstructing trajectories from scratch and the property of pure-pursuit method to compute the perceptual distance based on speed allow our method to adapt to different speed scenarios. In addition, we incorporate the real state of the future trajectory in the reference trajectory, when it enters the perceptual range, our generated adversarial trajectory can be naturally biased towards the future real state, which ensures the smoothness of the junction between the adversarial trajectory and the future trajectory.


\section{EXPERIMENTS}

\begin{table*}[t]
\caption{Average prediction error before and after adversarial attack.}
\centering
\label{Table.result1}
\begin{tabular}{|c|c|c|c|c|c|}
\hline
\multirow{2}{*}{Model}                  & \multirow{2}{*}{Dataset}         & ADE           & FDE                                & MR                                 & ORR                                \\ \cline{3-6} 
                                        &                                  & Normal/Attack & \multicolumn{1}{l|}{Normal/Attack} & \multicolumn{1}{l|}{Normal/Attack} & \multicolumn{1}{l|}{Normal/Attack} \\ \hline
\multirow{2}{*}{Grip++}                   & nuScenes                         & 4.34/7.43     & 10.39/16.04                        & 21\%/42\%                          & 3.7\%/8.5\%                        \\ \cline{2-6} 
                                        & \multicolumn{1}{l|}{Apolloscape} & 1.66/3.93     & 3.18/7.01                          & 2\%/26\%                           & 0.3\%/5.3\%                        \\ \hline
\multirow{2}{*}{Trajectron++}           & nuScenes                         & 3.84/7.31     & 11.26/17.92                        & 19\%/56\%                          & 1.6\%/8.6\%                        \\ \cline{2-6} 
                                        & \multicolumn{1}{l|}{Apolloscape} & 1.00/4.28     & 2.24/7.37                          & 0\%/28\%                           & 0.0\%/8.1\%                        \\ \hline
\multicolumn{1}{|l|}{Trajectron++(map)} & nuScenes                         & 1.99/5.34     & 5.38/10.83                         & 2\%/9\%                            & 0.3\%/2.2\%                        \\ \hline
\end{tabular}
\end{table*}

\begin{figure*}[!t]
\centering
\includegraphics[width=1.0\linewidth]{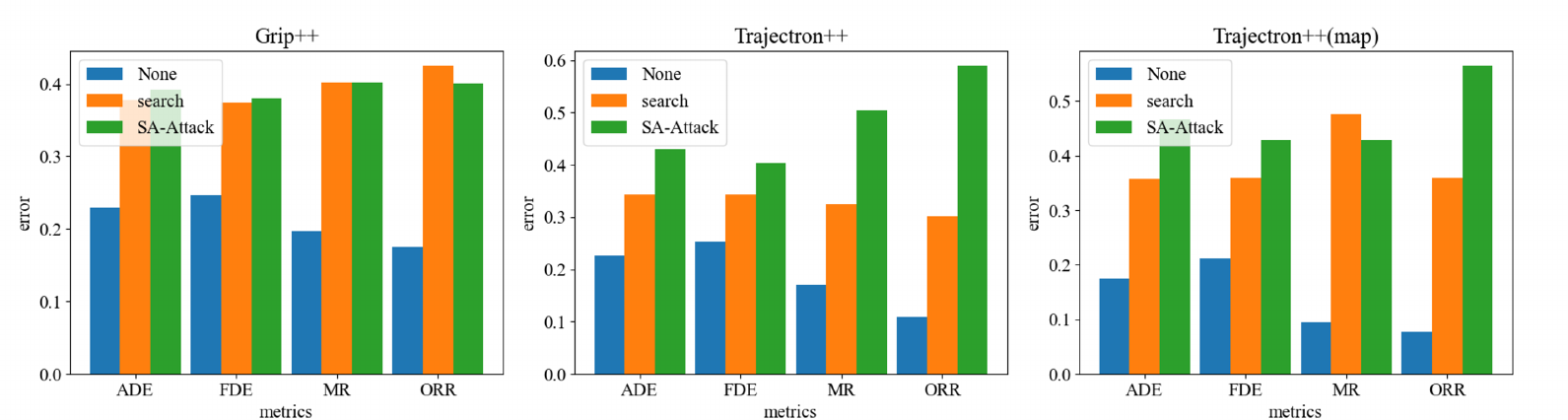}
\caption{Comparison of different attack methods.}
\label{fig:method_compare}
\end{figure*}

\subsection{\textbf{Experiment Set-up}}

\textbf{Datasets}: In our implementation, we consider two datasets, nuScenes\cite{caesar2020nuscenes} and Apolloscape\cite{huang2018apolloscape}, they both contain a large-scale data covering multiple cities, multiple weather and diverse traffic scenarios, and have the same sampling frequency of $2$Hz. According to the official recommendations, for the nuScenes, we select history trajectory length $L_{I}=4$, and future trajectory length $L_{O}=12$. For the Apolloscape dataset, we select history trajectory length $L_{I}=6$, and future trajectory length $L_{O}=6$. One hundred scenarios are randomly selected in each dataset.

\textbf{Models}: We consider two state-of-the-art trajectory prediction models, Trajectron++\cite{salzmann2020trajectron++} and Grip++\cite{li2019grip++}. We evaluate the results of these two models on nuScenes~\cite{caesar2020nuscenes} and Apolloscape~\cite{huang2018apolloscape}, respectively. By combining semantic maps, we performed additional evaluations on the nuScenes dataset using Trajectron++, this version is notated as Trajectron++ (map). In addition, Trajectron++ can be used to generate multimodal trajectory prediction, in which we select the predicted trajectory with the highest probability as the final result.

\textbf{Attack methods}: We select the search-based attack proposed by Zhang et al.\cite{zhang2022adversarial} as the baseline, henceforth referred to as \textit{search}. To make fair comparisons, we re-implement \textit{search}. For both SA-Attack and \textit{search}, the perturbations are optimised using the Adam optimiser, the learning rate of the optimisation process is set to 0.01 and the maximum number of iterations is 50.

\textbf{Metrics}:
Since there is no binary judgement of attack success for the trajectory prediction task, we evaluate it using a quantitative prediction error, which contains the following four metrics: 

Average Displacement Error (ADE): Mean $\ell_{2}$ distance between the ground truth and predicted trajectories.

Final Displacement Error (FDE): $\ell_{2}$ distance between the predicted final position and the ground truth final position at the prediction horizon $L_{O}$.

Miss Rates (MR): Number of trajectories not correctly predicted divided by the overall number of trajectories.

Off Road Rates (ORR): Number of off-road trajectories divided by number of overall trajectories.

\textbf{Implementation details}:

During the reference trajectory generation process, we generate 20 sets of randomly initialised perturbations. During the feasible trajectory reconstruction, we empirically set the perceptual distance $p_{v}$ to be an integer multiple of the scenario's average speed $\alpha =2$, and the length of each clothoid arc is 0.2m. For simplicity, we set the boundary of
the perturbation to a fixed value of $1m$, and the learning rate is fixed.

\subsection{\textbf{Main Results}}

\textbf{Trajectory prediction under attacks}:
First, for each combination of model and dataset, we analyse the effectiveness of SA-Attack. We present the average prediction errors before and after the perturbation in Table \ref{Table.result1}. SA-Attack increases the average prediction error ADE/FDE of trajectory predictions compared to normal predictions by $120\%/82\%$. In addition, we leverage MR and ORR to further describe the ability of the model to predict real trajectory. Through experiments, we demonstrate that SA-Attack is effective with different datasets and different models. From experimental results, we also find that Trajectron++ has higher prediction accuracy compared to the Grip++, which is attributed to the heterogeneous data integrated by Trajectron++. By encoding the map information, the ADE/FDE of the Trajectron++ decreases by $48\%/52\%$, respectively. However, high accuracy does not mean high robustness, Trajectron++ has higher attack sensitivity compared to the Grip++. From Table \ref{Table.result1}, we can find that SA-Attack presents a greater impact on the Trajectron++ model compared to Grip++.

\begin{figure}[!t]
\centering
\includegraphics[width=1.0\linewidth]{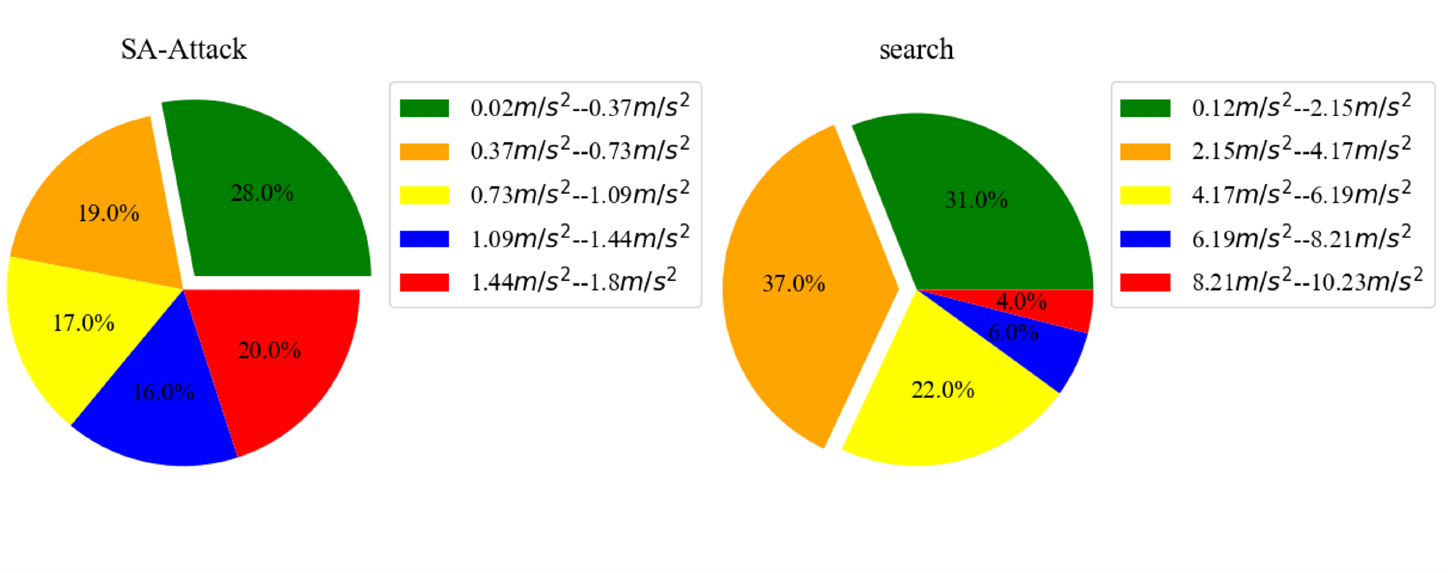}
\caption{Quantitative comparison of adversarial trajectories at maximum acceleration.}
\label{fig:velocity variation}
\end{figure}

\begin{figure*}[t]
\centering
\includegraphics[width=1.0\textwidth]{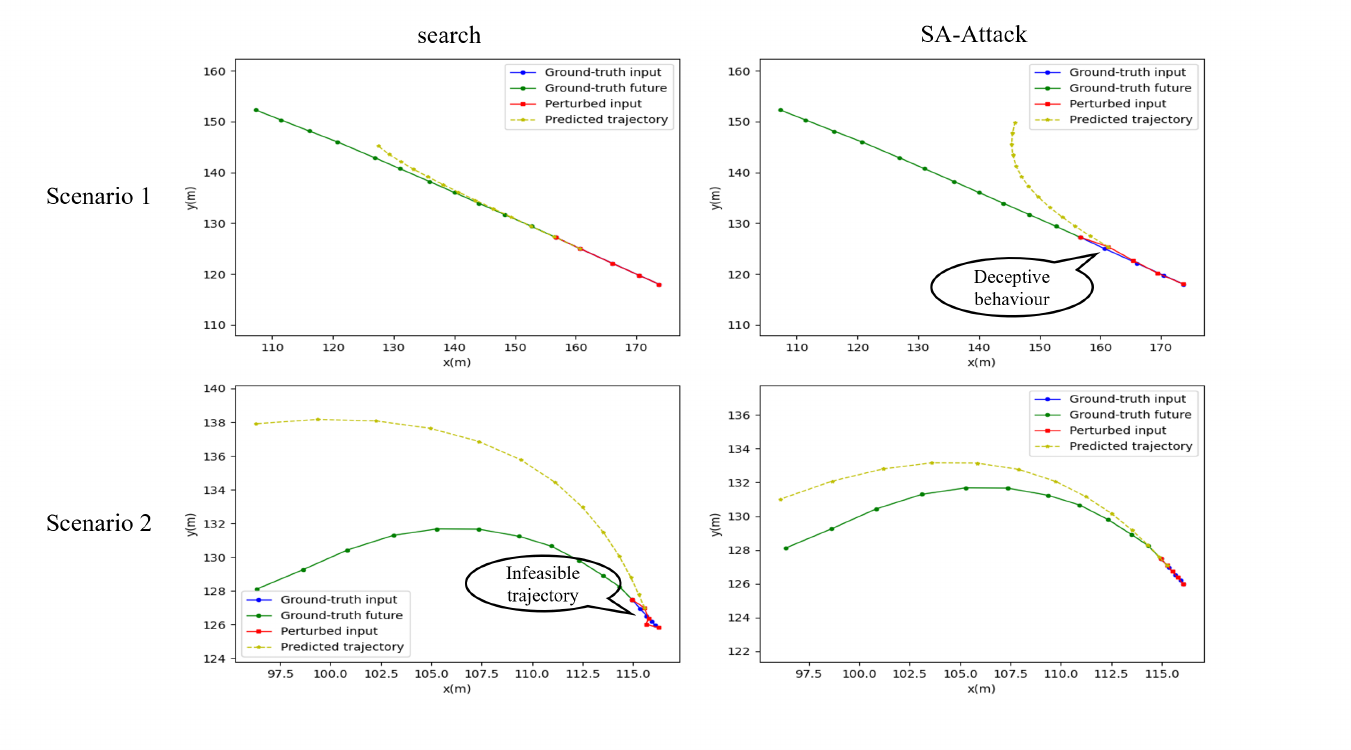}
\caption{Qualitative comparison of the feasibility of adversarial trajectories.}
\label{fig:Qualitative}
\end{figure*}

Next, we compare the effectiveness of the SA-Attack and \textit{search} method~\cite{zhang2022adversarial} on the nuScenes dataset. In order to better show the comparison of attack effects, we map all metrics to a range of 0-1 and present the results via histograms, as shown in Fig
\ref{fig:method_compare}. We denote the unattacked trajectory prediction model as None. Both SA-Attack and \textit{search}~\cite{zhang2022adversarial} produce effective attacks on the trajectory prediction model. SA-Attack outperforms \textit{search}~\cite{zhang2022adversarial} on both ADE and FDE metrics. In addition, SA-Attack achieves similar results to \textit{search}~\cite{zhang2022adversarial} on the Grip++ model. However, SA-Attack is significantly more effective than \textit{search}~\cite{zhang2022adversarial} on both the Trajectron++ model and the Trajectron++ (map) model.

\textbf{Adversarial trajectory feasibility:} We analyze the feasibility of generating adversarial trajectories for SA-Attack and \textit{search}~\cite{zhang2022adversarial} qualitatively and quantitatively. In Fig \ref{fig:velocity variation}, we use the maximum acceleration to represent the difficulty of reproducing the trajectory. Around 28\% of the adversarial trajectories produced by SA-Attack are within the range of $0.02m/s^2-0.37m/s^2$, and the maximum acceleration is within $1.8m/s^2$ for all scenarios. Around 37\% of the adversarial trajectories produced by \textit{search}\cite{zhang2022adversarial} are within the range $2.15m/s^2-4.17m/s^2$, and reproducing these adversarial trajectories requires larger braking behaviours, such as sharp braking. The maximum acceleration of an ordinary car is often within $6m/s^2$. About 10\% of the adversarial trajectories produced by \textit{search}~\cite{zhang2022adversarial} exceed this range, and these adversarial trajectories are likely to be infeasible and easily identified as anomalous trajectories. As a result, the adversarial trajectories generated by SA-Attack are more realistic and feasible.

In Fig. \ref{fig:Qualitative}, we visualise the adversarial trajectories generated by \textit{search}\cite{zhang2022adversarial} and SA-Attack method. We demonstrate that our method is able to generate adversarial trajectories that are both realistic and effective. In scenario $1$, our method generates effective attack through tiny deceptive behaviour and the adversarial trajectories can be smoothly connected to future trajectory. However, \textit{search}~\cite{zhang2022adversarial} does not produce an effective attack. In scenario $2$, our method demonstrates that it is also well adapted to the low-speed trajectory. In contrast, \textit{search}\cite{zhang2022adversarial} produces an unsmooth, infeasible trajectory in this scenario. Such a trajectory can be easily detected and defined as the anomalous trajectory, which limits its scalability.

\subsection{\textbf{Discussion}}
Based on the above experiment results, several findings can be highlighted.
\begin{itemize}
\item The high accuracy of trajectory prediction models does not mean that they are also robust. Under our stealth attack, the performance of the trajectory prediction models meets a huge degradation, which raises serious security issues. In our experiments, Trajectron++~\cite{salzmann2020trajectron++} is more vulnerable compared to Grip++. It is possible that Trajectron++ improves the prediction accuracy of the model on the dataset by integrating heterogeneous data, but it also reduces the robustness of the model, making it more susceptible to adversarial attacks.
\item Our proposed SA-Attack method achieves better attack performance compared to \textit{search}\cite{zhang2022adversarial}. It suggests that a direct search for the maximum perturbation under constraints tends to get stuck in the local optima and fails to generate successful adversarial trajectories. Instead, our method first explores the sensitive region of the model under unconstrained conditions and later reconstructs the feasible trajectories, which can effectively avoid local optima and produce better attack results.
\item We analyze the speed properties of different attack methods for generating adversarial trajectories. The \textit{search} method~\cite{zhang2022adversarial} generates adversarial trajectories with a larger range of speed variations, deceiving the trajectory prediction model by emergency braking, e.g., sharp acceleration, sharp deceleration. However, such adversarial trajectories can be easily detected as abnormal trajectories. Our method produces adversarial trajectories with a smaller range of speed variations, generating a more effective attack through polite driving behaviors with high attack stealthiness.
\item We visualize the adversarial trajectories generated by different attacks. The \textit{search} method~\cite{zhang2022adversarial} generates adversarial trajectories by constraining certain physical properties within a fixed range, which is not suitable to adapt to the high-speed and low-speed scenarios, while our method effectively adapts to different-speed scenarios by reconstructing the adversarial trajectories from scratch using the pure pursuit strategy.
\item Our proposed SA-Attack method needs to access the model parameters, which limits its real applications. In the future, several research directions will be explorable. First, the proposal of novel defense methods will be significant in enhancing
the robustness of trajectory prediction models. Second,
the black-box attacks for trajectory prediction are more
practical in real applications. Last but not least, effective
real scenario reconstruction can contribute to a better
understanding of the robustness of the trajectory prediction model.

\end{itemize}

\section{CONCLUSIONS}

This paper proposes a new adversarial attack method named SA-Attack for malicious interference on the trajectory prediction task. We guarantee the smoothness of the generated adversarial trajectories using the continuous curvature model, and we improve the adaptability of our attack method to scenarios with different speeds using the vehicle-following method. The experimental results demonstrate the effectiveness and stealthiness of our proposed attack method. We confirm that our proposed method can evaluate and enhance the robustness of trajectory prediction models based on deep learning.  

\section*{ACKNOWLEDGMENT}

This work was supported by the Shanghai International Science and Technology Cooperation Project No.22510712000 and the Special Funds of the Tongji University for ``Sino-German Cooperation 2.0 Strategy" No. ZD2023001. The authors would like to thank TÜV SÜD for the kind and generous support. We are also grateful for the efforts from our colleagues in Sino German Center of Intelligent Systems.


\bibliographystyle{unsrt}
\bibliography{root}

\begin{thebibliography}{10}

\bibitem{yuan2021agentformer}
Ye~Yuan, Xinshuo Weng, Yanglan Ou, and Kris~M Kitani.
\newblock Agentformer: Agent-aware transformers for socio-temporal multi-agent forecasting.
\newblock In {\em Proceedings of the IEEE/CVF International Conference on Computer Vision}, pages 9813--9823, 2021.

\bibitem{yin2021memory}
Huilin Yin, Jie Wang, Jia Lin, Daguang Han, Chunli Ying, and Qian Meng.
\newblock A memory-attention hierarchical model for driving-behavior recognition and motion prediction.
\newblock {\em International journal of automotive technology}, 22:895--908, 2021.

\bibitem{salzmann2020trajectron++}
Tim Salzmann, Boris Ivanovic, Punarjay Chakravarty, and Marco Pavone.
\newblock Trajectron++: Dynamically-feasible trajectory forecasting with heterogeneous data.
\newblock In {\em Computer Vision--ECCV 2020: 16th European Conference, Glasgow, UK, August 23--28, 2020, Proceedings, Part XVIII 16}, pages 683--700. Springer, 2020.

\bibitem{li2019grip++}
Xin Li, Xiaowen Ying, and Mooi~Choo Chuah.
\newblock Grip++: Enhanced graph-based interaction-aware trajectory prediction for autonomous driving.
\newblock {\em arXiv preprint arXiv:1907.07792}, 2019.

\bibitem{cao2022advdo}
Yulong Cao, Chaowei Xiao, Anima Anandkumar, Danfei Xu, and Marco Pavone.
\newblock Advdo: Realistic adversarial attacks for trajectory prediction.
\newblock In {\em European Conference on Computer Vision}, pages 36--52. Springer, 2022.

\bibitem{zhang2022adversarial}
Qingzhao Zhang, Shengtuo Hu, Jiachen Sun, Qi~Alfred Chen, and Z~Morley Mao.
\newblock On adversarial robustness of trajectory prediction for autonomous vehicles.
\newblock In {\em Proceedings of the IEEE/CVF Conference on Computer Vision and Pattern Recognition}, pages 15159--15168, 2022.

\bibitem{wang2021human}
Jiahang Wang, Sheng Jin, Wentao Liu, Weizhong Liu, Chen Qian, and Ping Luo.
\newblock When human pose estimation meets robustness: Adversarial algorithms and benchmarks.
\newblock In {\em Proceedings of the IEEE/CVF conference on computer vision and pattern recognition}, pages 11855--11864, 2021.

\bibitem{pivtoraiko2011kinodynamic}
Mihail Pivtoraiko and Alonzo Kelly.
\newblock Kinodynamic motion planning with state lattice motion primitives.
\newblock In {\em 2011 IEEE/RSJ International Conference on Intelligent Robots and Systems}, pages 2172--2179. IEEE, 2011.

\bibitem{sprunk2011online}
Christoph Sprunk, Boris Lau, Patrick Pfaffz, and Wolfram Burgard.
\newblock Online generation of kinodynamic trajectories for non-circular omnidirectional robots.
\newblock In {\em 2011 IEEE International Conference on Robotics and Automation}, pages 72--77. IEEE, 2011.

\bibitem{andreasson2015fast}
Henrik Andreasson, Jari Saarinen, Marcello Cirillo, Todor Stoyanov, and Achim~J Lilienthal.
\newblock Fast, continuous state path smoothing to improve navigation accuracy.
\newblock In {\em 2015 IEEE International Conference on Robotics and Automation (ICRA)}, pages 662--669. IEEE, 2015.

\bibitem{zhao2021tnt}
Hang Zhao, Jiyang Gao, Tian Lan, Chen Sun, Ben Sapp, Balakrishnan Varadarajan, Yue Shen, Yi~Shen, Yuning Chai, Cordelia Schmid, et~al.
\newblock Tnt: Target-driven trajectory prediction.
\newblock In {\em Conference on Robot Learning}, pages 895--904. PMLR, 2021.

\bibitem{caesar2020nuscenes}
Holger Caesar, Varun Bankiti, Alex~H Lang, Sourabh Vora, Venice~Erin Liong, Qiang Xu, Anush Krishnan, Yu~Pan, Giancarlo Baldan, and Oscar Beijbom.
\newblock nuscenes: A multimodal dataset for autonomous driving.
\newblock In {\em Proceedings of the IEEE/CVF conference on computer vision and pattern recognition}, pages 11621--11631, 2020.

\bibitem{huang2018apolloscape}
Xinyu Huang, Xinjing Cheng, Qichuan Geng, Binbin Cao, Dingfu Zhou, Peng Wang, Yuanqing Lin, and Ruigang Yang.
\newblock The apolloscape dataset for autonomous driving.
\newblock In {\em Proceedings of the IEEE conference on computer vision and pattern recognition workshops}, pages 954--960, 2018.

\bibitem{alahi2016social}
Alexandre Alahi, Kratarth Goel, Vignesh Ramanathan, Alexandre Robicquet, Li~Fei-Fei, and Silvio Savarese.
\newblock Social lstm: Human trajectory prediction in crowded spaces.
\newblock In {\em Proceedings of the IEEE conference on computer vision and pattern recognition}, pages 961--971, 2016.

\bibitem{vemula2018social}
Anirudh Vemula, Katharina Muelling, and Jean Oh.
\newblock Social attention: Modeling attention in human crowds.
\newblock In {\em 2018 IEEE international Conference on Robotics and Automation (ICRA)}, pages 4601--4607. IEEE, 2018.

\bibitem{yan2020trajectory}
Jun Yan, Zifeng Peng, Huilin Yin, Jie Wang, Xiao Wang, Yuesong Shen, Walter Stechele, and Daniel Cremers.
\newblock Trajectory prediction for intelligent vehicles using spatial-attention mechanism.
\newblock {\em IET Intelligent Transport Systems}, 14(13):1855--1863, 2020.

\bibitem{gao2020vectornet}
Jiyang Gao, Chen Sun, Hang Zhao, Yi~Shen, Dragomir Anguelov, Congcong Li, and Cordelia Schmid.
\newblock Vectornet: Encoding hd maps and agent dynamics from vectorized representation.
\newblock In {\em Proceedings of the IEEE/CVF Conference on Computer Vision and Pattern Recognition}, pages 11525--11533, 2020.

\bibitem{zhou2022hivt}
Zikang Zhou, Luyao Ye, Jianping Wang, Kui Wu, and Kejie Lu.
\newblock Hivt: Hierarchical vector transformer for multi-agent motion prediction.
\newblock In {\em Proceedings of the IEEE/CVF Conference on Computer Vision and Pattern Recognition}, pages 8823--8833, 2022.

\bibitem{zhang2019graph}
Si~Zhang, Hanghang Tong, Jiejun Xu, and Ross Maciejewski.
\newblock Graph convolutional networks: a comprehensive review.
\newblock {\em Computational Social Networks}, 6(1):1--23, 2019.

\bibitem{cui2019multimodal}
Henggang Cui, Vladan Radosavljevic, Fang-Chieh Chou, Tsung-Han Lin, Thi Nguyen, Tzu-Kuo Huang, Jeff Schneider, and Nemanja Djuric.
\newblock Multimodal trajectory predictions for autonomous driving using deep convolutional networks.
\newblock In {\em 2019 International Conference on Robotics and Automation (ICRA)}, pages 2090--2096. IEEE, 2019.

\bibitem{ma2019trafficpredict}
Yuexin Ma, Xinge Zhu, Sibo Zhang, Ruigang Yang, Wenping Wang, and Dinesh Manocha.
\newblock Trafficpredict: Trajectory prediction for heterogeneous traffic-agents.
\newblock In {\em Proceedings of the AAAI conference on artificial intelligence}, volume~33, pages 6120--6127, 2019.

\bibitem{zhang2019towards}
Haichao Zhang and Jianyu Wang.
\newblock Towards adversarially robust object detection.
\newblock In {\em Proceedings of the IEEE/CVF International Conference on Computer Vision}, pages 421--430, 2019.

\bibitem{yan2023adversarial}
Jun Yan, Huilin Yin, Bin Ye, Wanchen Ge, Hao Zhang, and Gerhard Rigoll.
\newblock An adversarial attack on salient regions of traffic sign.
\newblock {\em Automotive Innovation}, pages 1--14, 2023.

\bibitem{jia2020fooling}
Yunhan~Jia Jia, Yantao Lu, Junjie Shen, Qi~Alfred Chen, Hao Chen, Zhenyu Zhong, and Tao~Wei Wei.
\newblock Fooling detection alone is not enough: Adversarial attack against multiple object tracking.
\newblock In {\em International Conference on Learning Representations (ICLR'20)}, 2020.

\bibitem{sato2021dirty}
Takami Sato, Junjie Shen, Ningfei Wang, Yunhan Jia, Xue Lin, and Qi~Alfred Chen.
\newblock Dirty road can attack: Security of deep learning based automated lane centering under $\{$Physical-World$\}$ attack.
\newblock In {\em 30th USENIX Security Symposium (USENIX Security 21)}, pages 3309--3326, 2021.

\bibitem{yin2022adversarial}
Huilin Yin, Ruining Wang, Boyu Liu, and Jun Yan.
\newblock On adversarial robustness of semantic segmentation models for automated driving.
\newblock In {\em 2022 IEEE Intelligent Vehicles Symposium (IV)}, pages 867--873. IEEE, 2022.

\bibitem{shannon1949communication}
Claude~E Shannon.
\newblock Communication theory of secrecy systems.
\newblock {\em The Bell system technical journal}, 28(4):656--715, 1949.

\bibitem{madry2017towards}
Aleksander Madry, Aleksandar Makelov, Ludwig Schmidt, Dimitris Tsipras, and Adrian Vladu.
\newblock Towards deep learning models resistant to adversarial attacks.
\newblock {\em arXiv preprint arXiv:1706.06083}, 2017.

\bibitem{snider2009automatic}
Jarrod~M Snider et~al.
\newblock Automatic steering methods for autonomous automobile path tracking.
\newblock {\em Robotics Institute, Pittsburgh, PA, Tech. Rep. CMU-RITR-09-08}, 2009.

\bibitem{fassbender2016motion}
Dennis Fassbender, Benjamin~C Heinrich, and Hans-Joachim Wuensche.
\newblock Motion planning for autonomous vehicles in highly constrained urban environments.
\newblock In {\em 2016 IEEE/RSJ international conference on intelligent robots and systems (IROS)}, pages 4708--4713. IEEE, 2016.

\end{thebibliography}

\end{document}